\documentclass[runningheads]{llncs}
\usepackage{xcolor}
\usepackage{graphicx}
\usepackage{url}

\usepackage{subcaption}
\usepackage{booktabs}
\usepackage{multirow}
\usepackage{siunitx}

\usepackage[misc,geometry]{ifsym}

\usepackage{amsmath}
\usepackage{amssymb}
\usepackage{algorithm}
\usepackage{algorithmic}

\usepackage{cleveref}
\begin{document}
\title{FedAC: An Adaptive Clustered Federated Learning Framework for Heterogeneous Data}
\titlerunning{FedAC: A Adaptive Clustered Federated Learning Framework}
%
\author{Yuxin Zhang\inst{1}
\and
Haoyu Chen\inst{1} \and Zheng Lin\inst{2} \and Zhe Chen \inst{1} \and
Jin Zhao \inst{1}\Letter}
\authorrunning{Y. Zhang et al.}
%
\institute{School of Computer Science, Fudan University, Shanghai 200438, China \email{yuxinzhang22@m.fudan.edu.cn} \and Department of Electrical and Electronic Engineering, The University of Hong Kong, Pok Fu Lam, Hong Kong.}
%
\maketitle              
\begin{abstract}
Clustered federated learning (CFL) is proposed to mitigate the performance deterioration stemming from data heterogeneity in federated learning (FL) by grouping similar clients for cluster-wise model training. However, current CFL methods struggle due to inadequate integration of global and intra-cluster knowledge and the absence of an efficient online model similarity metric, while treating the cluster count as a fixed hyperparameter limits flexibility and robustness. In this paper, we propose an adaptive CFL framework, named \textsf{FedAC}, which (1) efficiently integrates global knowledge into intra-cluster learning by decoupling neural networks and utilizing distinct aggregation methods for each submodule, significantly enhancing performance; (2) includes a cost-effective online model similarity metric based on dimensionality reduction; (3) incorporates a cluster number fine-tuning module for improved adaptability and scalability in complex, heterogeneous environments. Extensive experiments show that \textsf{FedAC} achieves superior empirical performance, increasing the test accuracy by around $1.82\%$ and $12.67\%$ on CIFAR-10 and CIFAR-100 datasets, respectively, under different non-IID settings compared to SOTA methods.

\keywords{Federated learning \and Multi-task learning \and Clustering \and Non-IID data \and Model similarity.}
\end{abstract}
\section{Introduction}

Machine learning (ML) has rapidly advanced, finding extensive applications across industries such as intelligent agent, autonomous driving and energy management \cite{lin2024efficient,distribute_neighborhood,zheng2023autofed,solar,lin2022channel,yuan2023graph,hu2023adaptive,lin2024adaptsfl}. Its progress is attributed to diverse datasets, enhancing generalization and reducing overfitting \cite{dataeffect1,hu2024collaborative}. However, in real-world scenarios, data may be distributed across mobile devices and the Internet of Things (IoT) \cite{combining}. Decentralized training of ML models becomes necessary because of privacy constraints \cite{lin2023pushing,dataprivacy} and network bandwidth limitations \cite{lyu2023optimal,chen2022multi,lin2023fedsn}, preventing the transmission of these data to a central server for centralized training. Fueled by this realistic need, federated learning (FL) \cite{fl1,FedAvg} was proposed, allowing multiple clients to collaboratively learn a global model without exchanging local data.

However, a significant practical challenge in FL lies in data heterogeneity. Clients may have non-IID data and diverse preferences, resulting in variations in the true risk at the local level, which is inconsistent with the existence of a global model suitable for all clients \cite{fedem,lin2023split}. Even when considering a macroscopic perspective of empirical risk to formulate the global objective, local updates to client models may deviate to varying degrees. This presents challenges in convergence, potentially leading to suboptimal outcomes as the average global model drifts away \cite{heter1}.


Clustered federated learning (CFL) \cite{CFL} employs multi-task learning \cite{multitask} to mitigate data heterogeneity by grouping clients into clusters with higher internal \textit{similarity} (more homogeneity). Despite extensive research, current CFL methods still have limitations: (1) These methods demonstrate isolated clusters, lacking the infusion of valuable global knowledge. Consequently, clients are unaware of potentially beneficial knowledge beyond their assigned clusters, resulting in overall suboptimal performance; (2) There is still a need for a computationally effective method to accurately measure the \textit{online} similarity between models for clustering adjustments throughout the entire training process; (3) Furthermore, current methods treat the cluster count ($K$) as a constant hyper-parameter throughout the process, overlooking the challenge of manually setting the optimal value in complex heterogeneous scenarios. For example, when training a predictive text model with CFL for millions of global smartphone users, the complex and interrelated hidden contextual factors like age, location, occupation, etc., makes it impossible to pre-set the optimal number of clusters. Hence, there is an urgent requirement for an adaptive approach to fine-tune and determine its optimal value during trainig.

Motivated by the above challenges, this paper proposes \textsf{FedAC}, an adaptive CFL framework, comprising the following main components: (1) Decoupling neural networks into submodules and employing distinct aggregation methods, \textsf{FedAC} effectively integrates global knowledge into clusters. This allows clients to learn from both cluster-specific and global dimensions simultaneously, ensuring optimal performance by striking a balance; (2) By integrating a cosine model similarity metric following dimensionality reduction, \textsf{FedAC} effectively captures the online similarity of models at a computationally economical cost; (3) \textsf{FedAC} incorporates a module that dynamically fine-tunes the cluster count based on the current clustering status (inter-cluster and intra-cluster model distances), eliminating the necessity of pre-specifying a fixed cluster count. Instead, it autonomously fine-tunes the count during training to discover the optimal value, thereby enhancing the system's robustness and flexibility. Extensive experiments demonstrate that \textsf{FedAC} outperforms SOTA methods in diverse heterogeneous scenarios. The effectiveness of each component is further validated through ablation experiments, showcasing the flexibility and robustness of \textsf{FedAC}.

The following summarizes our contributions.

\begin{itemize}
  \item In this paper, we propse \textsf{FedAC}, an efficient and adaptive CFL framework designed to tackle complex non-IID scenarios within FL. To the best of our knowledge, \textsf{FedAC} is the first CFL framework to achieve outstanding performance by introducing global knowledge for intra-cluster learning through the decoupling of neural networks.
  \item We develope an approach to online assessment of model similarity using dimensionality-reduced models. This allows the server to efficiently evaluate similarities among data-heterogeneous clients in a cost-effective manner, enhancing clustering effectiveness and system scalability.
  \item We present a design approach and implementation for adaptive adjustment of the total cluster count based on clustering status, enhancing the framework's flexibility and robustness in addressing the challenge of manually setting it in complex, heterogeneous scenarios.
  \item We performed experiments on diverse datasets in complex heterogeneous scenarios to demonstrate the outstanding overall performance and adaptability of \textsf{FedAC}, surpassing SOTA methods.
\end{itemize}

The rest of this paper is organized as follows. In \cref{sec:related work}, we review Heterogeneous FL and CFL approaches addressing the data heterogeneity issue. \cref{third:setting} formulates the non-IID problem in CFL and outlines the optimization goal. \cref{forth:methodology} introduces the framework and essential components of \textsf{FedAC}. Performance evaluations follow in \cref{fifth:exp}, and the paper concludes in \cref{sixth:conclu}.

\section{Related Work}
\label{sec:related work}

\subsection{Heterogeneous Federated Learning}

The performance of FL is significantly affected by data heterogeneity, specifically, the non-identically and independently distributed (non-IID) nature of the data. The differences in data distribution among each client can lead to biased model updates, and directly averaging local updates (as in \textsf{FedAvg} \cite{FedAvg}) may detrimentally impact the overall performance of the global model. \textsf{FedPer} \cite{FedPer} decouples models and aggregates them using diverse strategies. \textsf{Per-FedAvg} \cite{Per-FedAvg} blends \textsf{FedAvg} with model-agnostic meta-learning (MAML) for personalized models via fine-tuning, and this concept is extended in \textsf{pFedMe} \cite{pFedMe} using Moreau envelopes. \textsf{pFedHN} \cite{pFedHN} employs a hypernetwork to generate parameters for the personalized model of each client, and \textsf{FedVF} \cite{FedVF} generates personalized models through a two-stage training process.

CFL, as another extensively researched heterogeneous FL method, addresses non-IID issues by assuming that clients can be partitioned into several clusters. During the training process, the server needs to learn the appropriate cluster assignment for each client and simultaneously improve the cluster center models. 
An overview of CFL methods is presented in the following section.

\subsection{Clustered Federated Learning}

In \cite{FLHC}, a hierarchical clustering framework is introduced for FL. It reduces computation and communication loads by using an agglomerative formulation. In \cite{CFL}, hierarchical clustering is used as a postprocessing step in FL. However, the recursive bipartitioning framework incurs high costs, limiting feasibility for large-scale settings. \textsf{FedGroup} \cite{FedGroup} utilizes a static client cluster methodology, initiating cold start protocols for new clients and adopting Euclidean distance of decomposed cosine similarity (EDC) for clustering. These methods cluster clients only once offline, and their performance heavily depends on the effectiveness of the clustering, exhibiting limited flexibility and robustness.

Online CFL methods address these issues by continuously re-clustering during training. In \textsf{FeSEM} \cite{FeSEM}, the expectation-maximization (EM) algorithm resolves the distance-based objective clustering issue by optimally pairing clients with their respective cluster centers. \cite{IFCA} introduced \textsf{IFCA}, which employs $K$ global models distributed to all clients for local loss computation to perform clustering, albeit with increased communication overhead. However, having clients directly adopt the cluster center model for inference compromises performance in complex environments. \textsf{CGPFL} \cite{CGPFL} remedies the problem by preserving personalized models for each client. The cluster center facilitates the learning of intra-cluster knowledge through soft regularization.

CFL methods face challenges: (1) they frequently confine knowledge sharing to particular clusters, lacking effective strategies to integrate global knowledge; (2) online CFL methods involve substantial computational overhead in re-clustering, lacking an efficient and cost-effective similarity metric; (3) all CFL methods require prior specification of the total cluster count, limiting scalability and flexibility in complex environments.

\section{Problem Formulation}
\label{third:setting}

In the context of FL, each client $i$ possesses its private dataset $\mathcal{D}_i$ from distribution $ \mathbb{P}_i(x, y)$, where $x$ and $y$ represent the input features and corresponding labels. In a vanilla setting (FedAvg), clients collectively contribute to a shared model $f(\omega;\cdot)$ parameterized by weights $\omega$. The objective function is:

\begin{equation}
\begin{split}
 \min_{\omega} :  \mathcal{F} =  \sum_{i=1}^m \frac{\left\vert \mathcal{D}_i \right\vert}{N} \mathcal{L}_i(\omega),
\end{split}
\end{equation}
where $\mathcal{L}_i(\omega) = \mathbb{E}_{(x,y)\sim \mathcal{D}_i} l(f(\omega;x);y)$ is the empirical
loss of client $i$, $m$ is the number of clients, and $N$ denotes the total number of instances over all clients. 

To further address data heterogeneity, personalized client-specific models, $ \{ f_i \}_{i \in [m]} $, are often employed across the system. While model architecture and hyper-parameters may differ among clients, in our study, we maintain identical model architectures for each, but with unique local parameters $ \{ \omega_i \}_{i \in [m]} $. Following this, clustering configurations' integration is proposed as the problem setting. The server retains 
$K$ cluster center models $\{\Omega_k\}_{k \in [K]}$, rather than a single global model, to steer local model updates:

\begin{equation}
\begin{split}
&\min_{\{ \omega_i \}} \frac{1}{m} \sum_{i=1}^m  \{ \mathcal{L}_i(\omega_i) + \frac{\mu}{2} \textrm{Regu}(\omega_i, \Omega_{k^*(i,R^*)}) \} ,\\
\textrm{s.t.} \quad &R^* \in \underset{R}{\arg\max} \frac{1}{m} \sum_{k=1}^K \sum_{i=1}^m R_{i,k} \textrm{Sim} (\omega_i, \Omega_k(\{ \omega_i \},R)),
\end{split}
\end{equation}
where $\mu$ is a regularization parameter, $R \in \mathbb{R}^{m \ast K}$ is the assignment matrix where $R_{i,k} = 1$ if client $i$ belongs to cluster $k$ else $R_{i,k} = 0$, $\Omega_k(\{ \omega_i \},R) = \frac{1}{\sum_{i=1}^m R_{i,k}}\sum_{i=1}^m R_{i,k} \omega_i$ is the central model of cluster $k$, and $k^*(i,R)$ represents the cluster to which client $i$ belongs, corresponding to the sole $1$ in client $i$'s row of the allocation matrix $R$. Note that $k^*$ is clearly defined per $R$'s definition. $\textrm{Regu}(\cdot,\cdot)$ regulates model difference, while $\textrm{Sim}(\cdot,\cdot)$ measures model similarity. Item $ \frac{\left\vert \mathcal{D}_i \right\vert}{N} $ is omitted from Eq. 1 to adopt a macroscopic view.

\section{Methodology}
\label{forth:methodology}

\subsection{Overview and Optimization Objective}

The proposed \textbf{Fed}erated Learning with \textbf{A}daptive \textbf{C}lustering (\textsf{FedAC}) addresses data heterogeneity and achieves efficient, adaptable client clustering. Firstly, to better integrate beneficial global knowledge into cluster, the model $f$ is decoupled into two submodules, exemplified by convolutional neural networks (convnets): the embedding $\phi$ (comprising shallow layers with convolutional modules) and the decision $h$ (comprising deep fully connected layers). Different aggregation strategies are designed for distinct submodules to represent global and intra-cluster knowledge, achieving optimal performance by balancing both aspects. Moreover, a cosine similarity metric based on dimensionality-reduced models is proposed for efficient online model similarity measurement to assist clustering throughout training. Additionally, a self-examining module evaluates the current cluster conditions and autonomously finetunes the cluster number. An overview of \textsf{FedAC}'s framework is shown in Fig. 1.

\begin{figure*}[h]
\centering
\includegraphics[width=.95\textwidth]{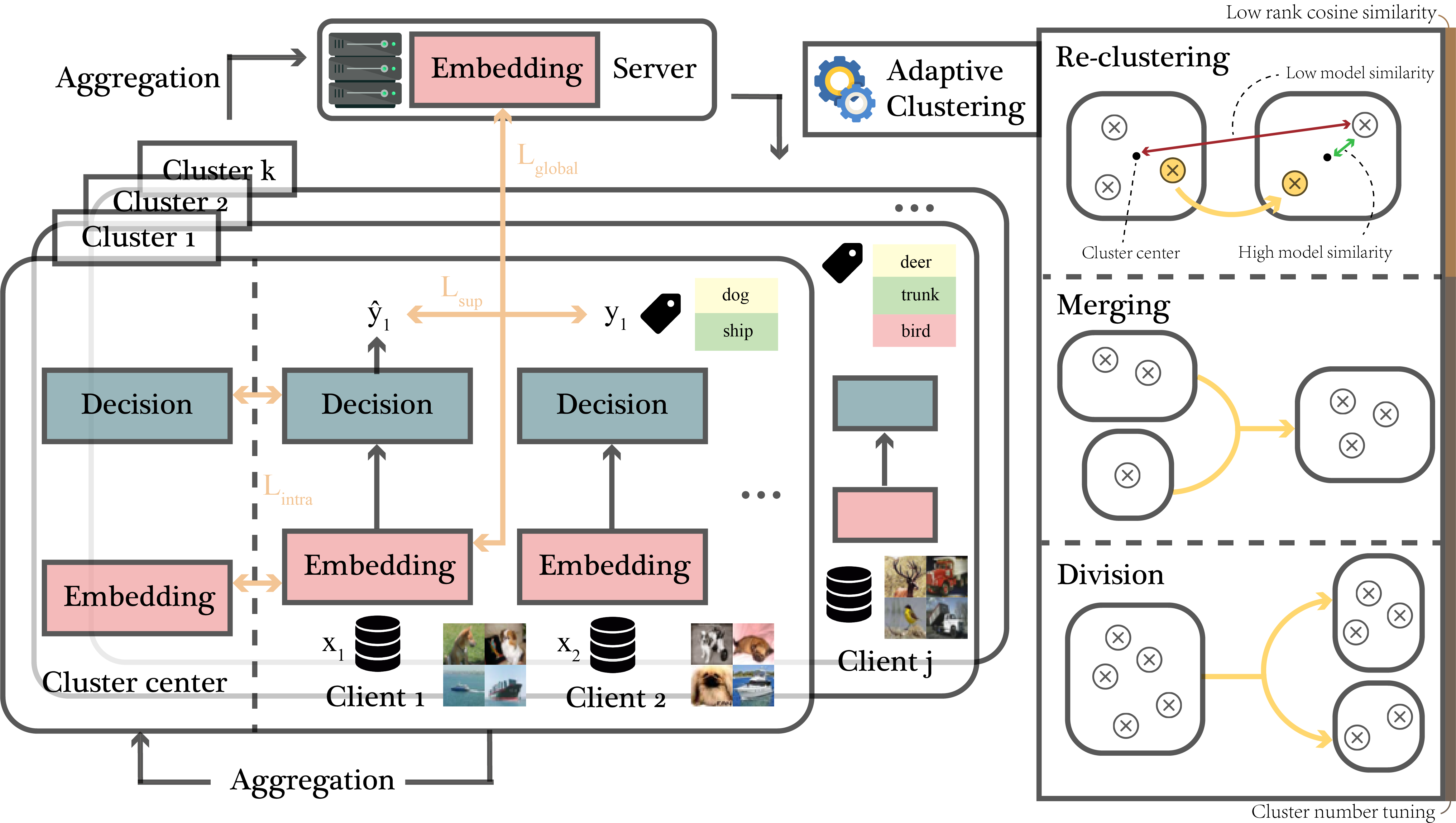}
\caption{The framwork of FedAC in the heterogeneous setting. The server clusters and aggregates clients to get a global embedding and multiple cluster center models. Clients update local models by minimizing the classification error loss ($L_{\text{sup}}$), intra-clustering regularization ($L_{\text{intra}}$), and global regularization ($L_{\text{global}}$).}
\label{framwork}
\end{figure*}\textbf{}

\begin{algorithm}[tb]
\caption{FedAC}
\label{alg:FedAC}
\textbf{Input}: learning rate $\eta$, hyper-parameters $\mu$ and $\lambda$, initial number of clusters $K$\\
\textbf{Output}: $ \{ \omega_i \}_{i \in [m]} $\\
\textbf{Server executes}:
\begin{algorithmic}[1] 
\STATE Initialize: $R^0$, $ \{ \omega_i^0 = (\phi_i^0, h_i^0)\}_{i \in [m]} $,  $ \{ \Omega_k^0 \}_{k \in [K]} $, $\Phi^{0}$
\FOR{each round $t = 0,1,...$}
\STATE Randomly selects a subset of clients $S_t$
\FOR{each client $i$ $\in$ $S_t$ \textbf{in parallel}}
\STATE Server sends  $\Omega^{t}_{k^*}$, $\Phi^{t}$ to client $i$
\STATE $ \omega_i^{t+1} \gets $ LocalUpdate ($ \omega_i^{t} $, $ \Omega^{t}_{k^*} $, $\Phi^{t}$, $\mu$, $\lambda$, $\eta$)
\STATE Clients $i$ sends $ \omega_i^{t+1} $ back
\ENDFOR
\STATE $ \Phi^{t+1} =\sum_{i \in S_t} \phi_i^{t+1} $
\STATE Calculate low-rank cosine model similarity \COMMENT{Algorithm 2}
\STATE $R^{t+1} \gets $ E-step ($ \{ \omega_i^{t+1} \}_{i \in [m]} $, $ \{ \Omega_k^t \}_{k \in [K]} $)
\STATE $ \{ \Omega_k^{t+1} \}_{k \in [K]} \gets $ M-step ($ \{ \omega_i^{t+1} \}_{i \in [m]} $, $ R^{t+1} $)
\STATE Cluster number tuning (CNT) \COMMENT{Algorithm 3}
\ENDFOR
\end{algorithmic}
\textbf{}

\textbf{LocalUpdate ($ \omega_i^{t}$, $\Omega^{t}_{k^*} $, $\Phi^{t}$, $\mu$, $\lambda$, $\eta$)}:

\begin{algorithmic}[1]
\STATE $\omega_i^{t,0} = \omega_i^{t}$
\FOR{each local epoch $r$ = 0,1,...}
\STATE Randomly selects a batch $\mathcal{B}_i$ from $\mathcal{D}_i$
\STATE  $\omega_i^{t,r+1} = \omega_i^{t,r} - \eta\nabla l_i(\omega_i^{t,r};\mathcal{B}_i) -\eta\mu(\omega_i^{t,r} - \Omega^{t}_{k^*}) - \eta\lambda(\phi_i^{t,r} - \Phi^t)$
\ENDFOR
\STATE \textbf{return} $\omega_i^{t,r+1}$
\end{algorithmic}
\end{algorithm}

We apply L2 distance and the proposed low-rank cosine model similarity, $LrCos$, into $\textrm{Regu}(\cdot,\cdot)$ and $\textrm{Sim}(\cdot,\cdot)$, respectively, within Eq. 2, defining the optimization objective of \textsf{FedAC}:

\begin{align}
\min_{\{ \omega_i \}} \frac{1}{m}\sum_{i=1}^m  \{ \underbrace{\mathcal{L}_i(\omega_i)}_{L_{\text{sup}}} + \frac{\mu}{2}\underbrace{|| \omega_i - \Omega_{k^*(i,R^*)} ||^2_2}_{L_{\text{intra}}}\nonumber + \frac{\lambda}{2}\underbrace{ || \phi_i - \Phi (\{\phi_i\})||^2_2}_{L_{\text{global}}}\} ,\nonumber
\end{align}
\begin{equation}
\textrm{s.t.} \quad R^* \in \underset{R}{\arg\max} \frac{1}{m} \sum_{k=1}^K \sum_{i=1}^m R_{i,k} LrCos (\omega_i, \Omega_k(\{ \omega_i \},R)),
\end{equation}
where hyper-parameter $\mu$ and $\lambda$ controls the level of regularization, and $\Phi(\{\phi_i\}) = \sum_{i=1}^m \phi_i$ is the global embedding. Note that the embedding regularization term has been included to capture global knowledge.

The bi-level optimization problem is addressed through alternate minimization, involving two iterative steps: (1) minimization w.r.t. $ \{ \omega_i \}_{i \in [m]} $ with $R$ fixed; and (2) minimization w.r.t. $R$ with $ \{ \omega_i \}_{i \in [m]} $ fixed. Step (1) requires clients to utilize local data for supervised learning, with regularization implemented through global embedding and the cluster center. In step (2), cluster assignments and center models are updated based on the $LrCos$ similarity, coupled with a EM-like algorithm. Additional details are presented in the subsequent sections, and Algorithm 1 provides the pseudocode.

\subsection{Integrating Global Knowledge into Clusters}

As one of the primary challenges in CFL methods is effectively integrating beneficial global knowledge into clusters, we employ a novel approach of decoupling neural networks for clients, allowing for finer control and balance between global and intra-cluster knowledge. This draws inspiration from the effective management of latent feature spaces in \cite{FedPer}, \cite{FedProto}, etc. In the context of convnets, shallow layers primarily handle pixel embedding and feature extraction tasks, capturing colors and edges from images. These features are considered advantageous for sharing among all clients, serving as the foundation for downstream tasks while being less affected by heterogeneous data distributions \cite{shareproto1}. The deep layers, which capture unique data distributions for each client, are preferred to be shared at the intra-cluster level, as they serve specific downstream tasks \cite{nosharedeep}.

Based on this, \textsf{FedAC} simultaneously updates global embedding $\Phi$ (line $9$ of Algorithm 1) and cluster-center models $\{ \Omega_k \}_{k \in [K]}$ (including embeddings and decisions, line $12$ of Algorithm 1) at each aggregation step. As client $i$ undergoes local updates, it learns global knowledge from $\Phi$, regulated by the term $L_{global}$ in Eq.3, while simultaneously learning intra-cluster knowledge from its corresponding cluster center model $\Omega_{k^*(i,R)}$, regulated by the term $L_{intra}$. By adjusting the corresponding regularization strengths of $L_{global}$ and $L_{intra}$ to balance their impacts on client's local updates, \textsf{FedAC} ensures the coordinated integration of intra-cluster and global knowledge, thereby achieving higher accuracy, as demonstrated in the experimental section.

\subsection{Low-rank Cosine Model Similarity}
\label{subsec:lrcos}

\begin{algorithm}[t]
\caption{Low-rank Cosine Similarity}
\label{alg:lrcos}

\textbf{Input}: $ \{ \omega_i \}_{i \in [m]} $, $ \{ \Omega_k \}_{k \in [K]} $\\
\textbf{Parameter}: reduced number of dimensions $D$\\
\textbf{Output}: $ LrCos_{i,j} $ between client $i$ and cluster $k$

\begin{algorithmic}[1] 
\STATE Mapping $ M \gets $ UpdateMap ($S_t$, $D$)  \COMMENT{Proceed sparsely}
\STATE $ LrCos_{i,j} = \frac{M \cdot \omega_i \cdot M \cdot \Omega_k}{||M \cdot \omega_i||_2||M \cdot \Omega_k||_2}  $
\end{algorithmic}
\textbf{UpdateMap ($S_t$, $D$)}:

\begin{algorithmic}[1]
\STATE $H \in \mathbb{R}^{dim(\omega) \times |S|} \gets [\omega_1,...,\omega_{|S|}]$
\STATE $ M \in \mathbb{R}^{D \times dim(\omega)} \gets $ PCA($H$, components=$D$)
\STATE \textbf{return} $M$
\end{algorithmic}
\end{algorithm}

Compared to one-shot clustering, online clustering methods, which continuously update model similarities and adjust clustering, often yield more effective clustering results but also come with increased computational overhead. We combine the benefits of one-shot and online clustering by sparsely computing the dimensionality reduction matrix and leveraging the cosine similarity of the reduced-dimensional model, namely, the low-rank cosine similarity $LrCos$, enabling efficient model similarity updates per round with low computational costs.

\begin{figure}[t]
\centering
\begin{subfigure}{0.24\textwidth}
\includegraphics[width=\textwidth]{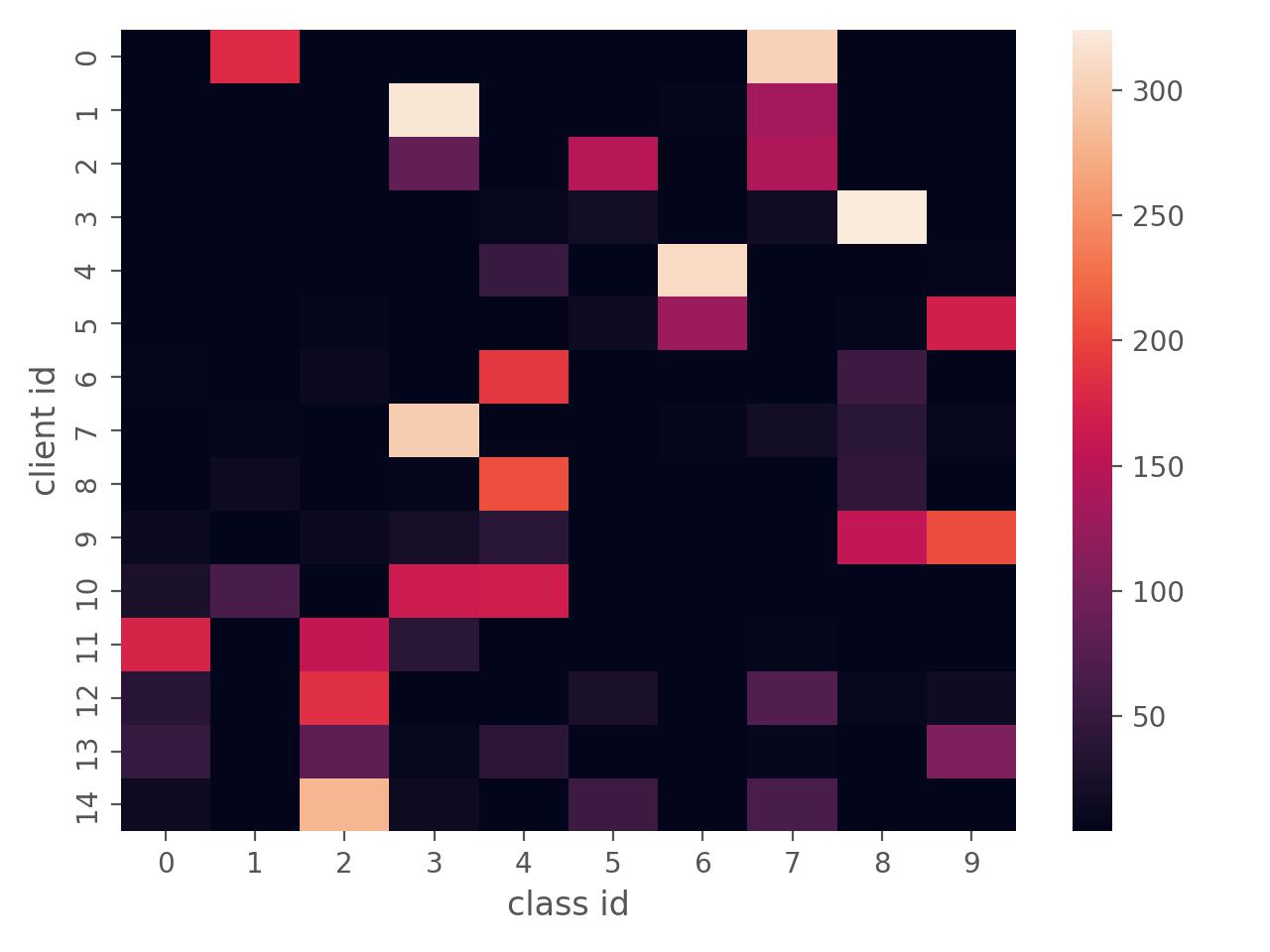}
    \caption{Data.}
    \label{fig:lrcos_data}
\end{subfigure}
\hfill
\begin{subfigure}{0.24\textwidth}
\includegraphics[width=\textwidth]{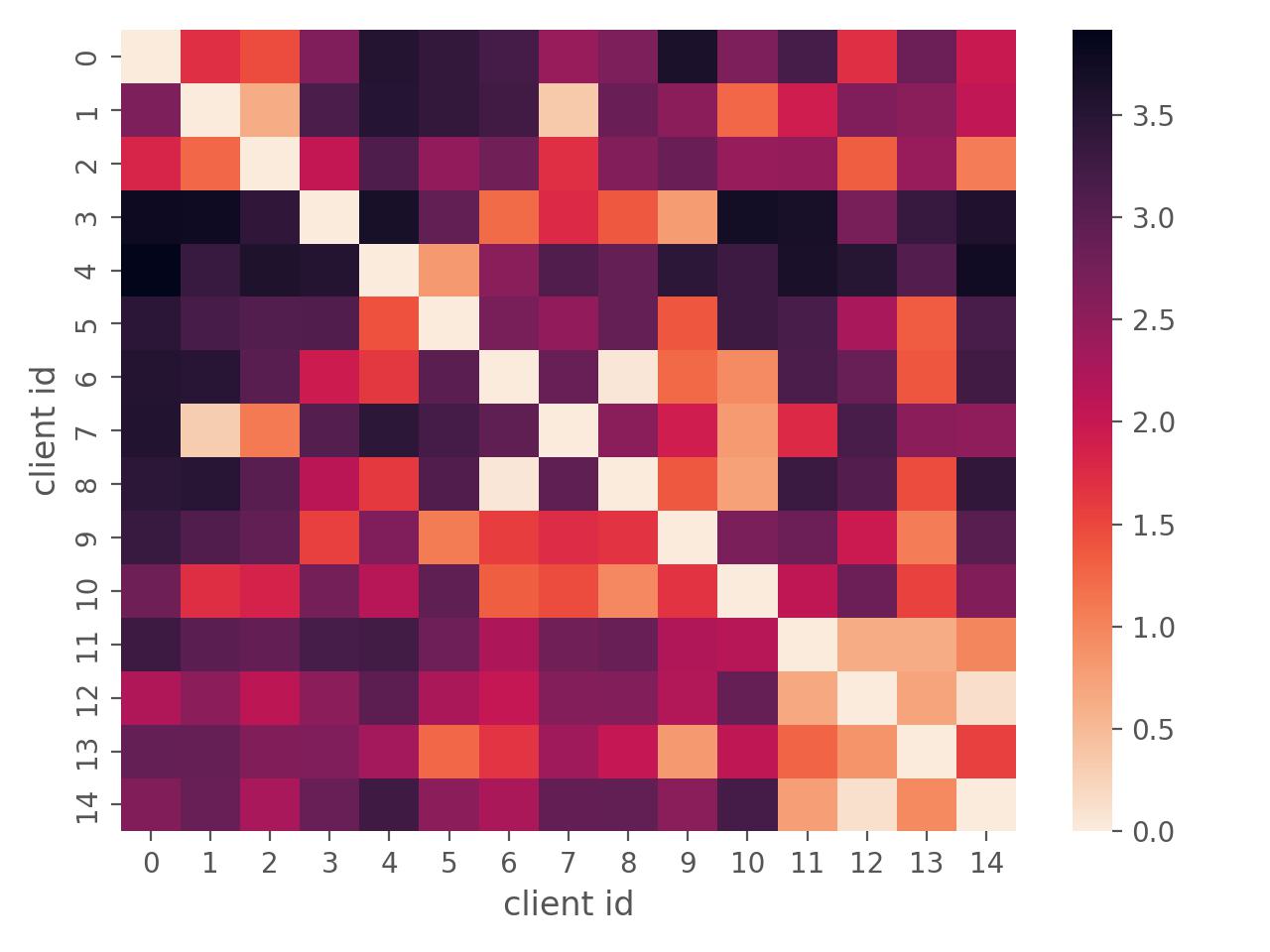}
    \caption{KL.}
    \label{fig:lrcos_KL}
\end{subfigure}
\hfill
\begin{subfigure}{0.24\textwidth}
\includegraphics[width=\textwidth]{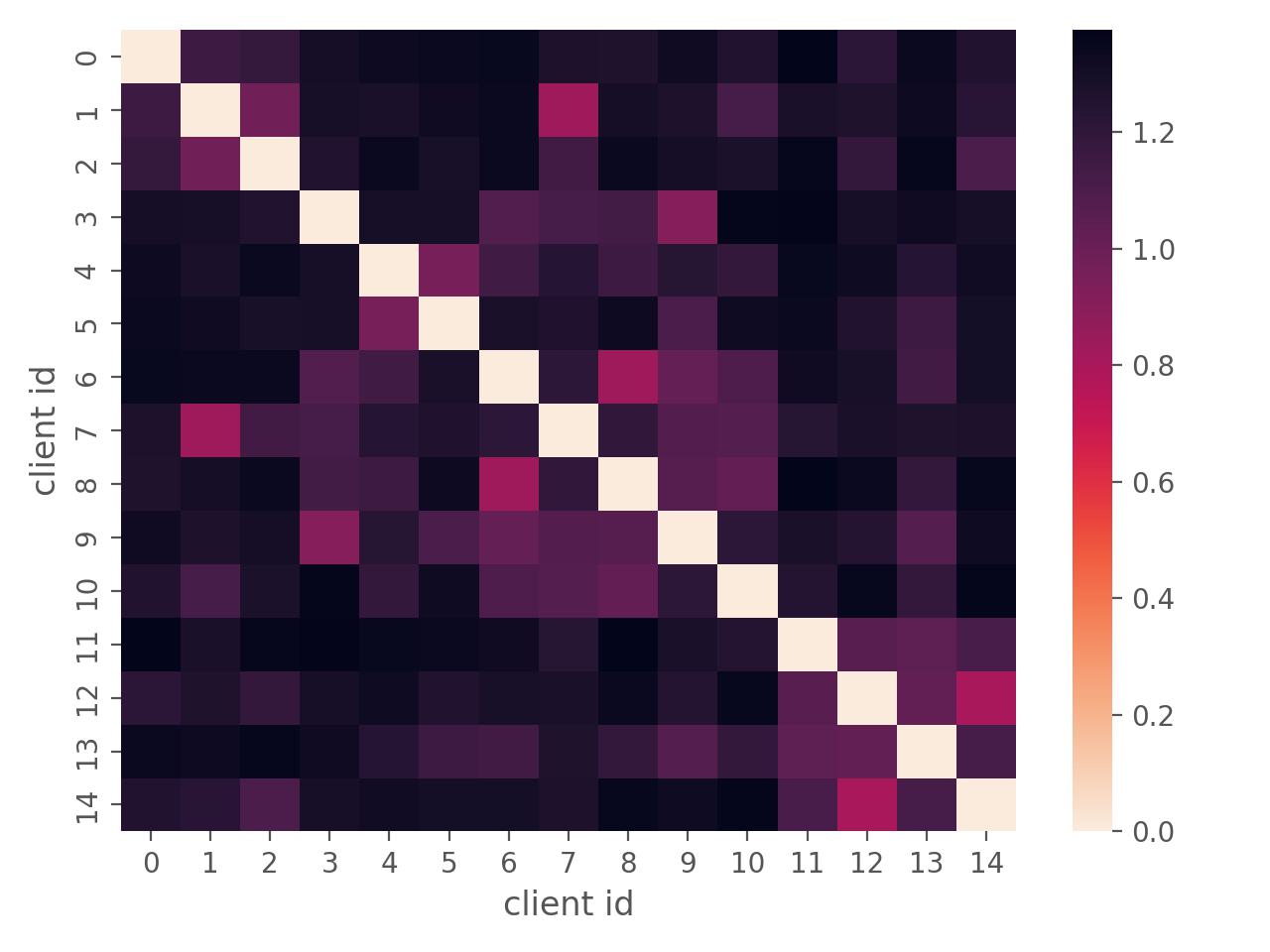}
    \caption{L2.}
    \label{fig:lrcos_L2}
\end{subfigure}
\hfill
\begin{subfigure}{0.24\textwidth}
    \includegraphics[width=\textwidth]{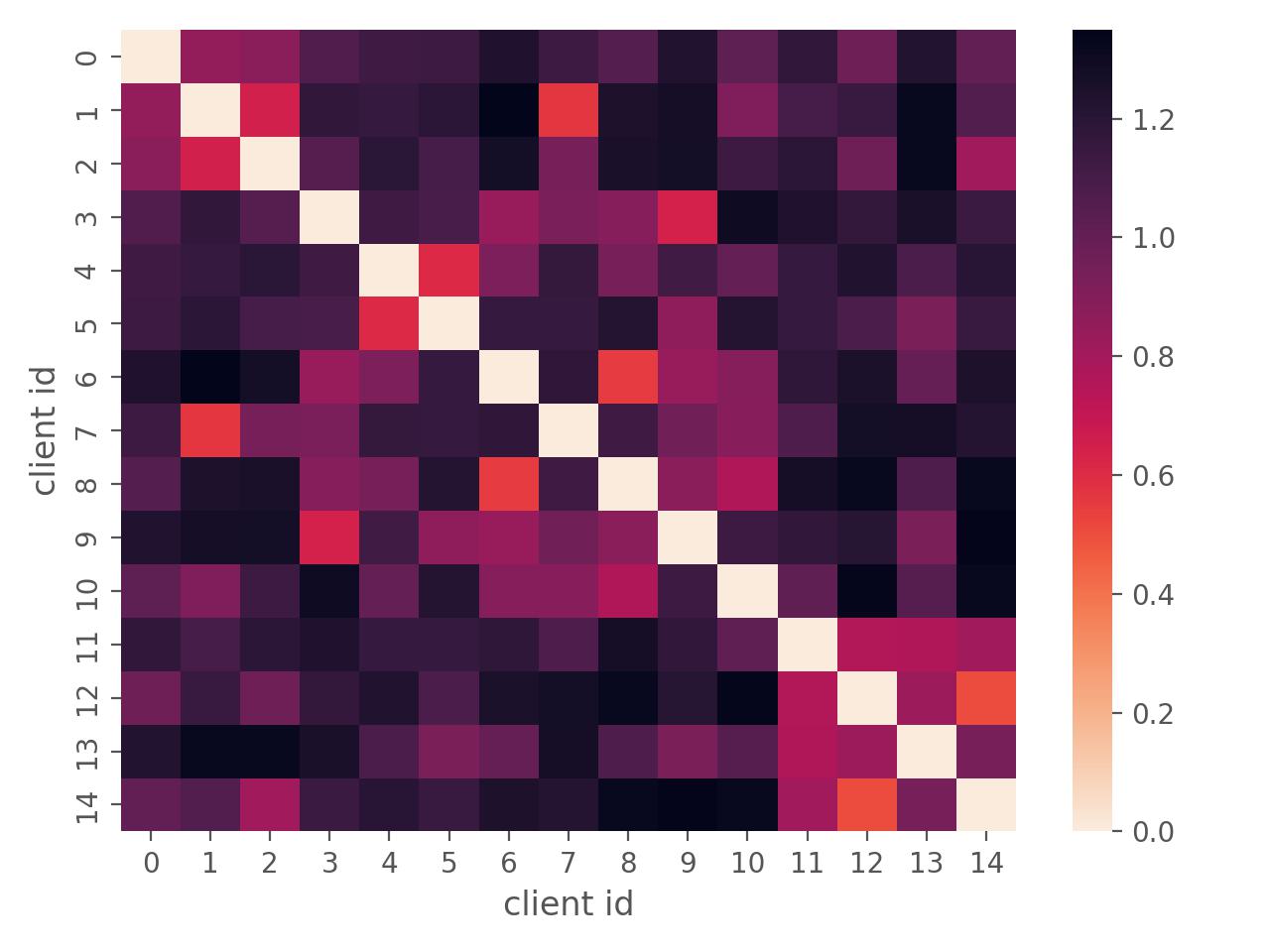}
    \caption{$LrCos$.}
    \label{fig:lrcos_LrCos}
\end{subfigure}
\caption{Clients employed FedAvg with CIFAR-10 \cite{CIFAR10} for $20$ epochs, succeeded by local fine-tuning, with model similarities subsequently evaluated.}
\label{fig:lrcosfigures}
\end{figure}

We utilize PCA \cite{PCA} for model dimensionality reduction, compatible with other methods. The server sporadically updates the dimensionality reduction matrix $M$ (e.g., every 100 rounds). Each client locally retains matrix $M$ and, while transmitting its model updates to the server, also sends the reduced model $M \cdot \omega$. After each update of the cluster center model $\Omega$, the server performs dimensionality reduction to acquire $M \cdot \Omega$. Following Algorithm 2, it uses cosine similarity to calculate the updated similarity between clients and cluster centers, enabling subsequent re-clustering.

Simulations confirm the effectiveness of $LrCos$. Fig. 2(a) illustrates the number of training samples held by each client for each class. Fig. 2(b) illustrates the Kullback-Leibler (KL) divergence between the data distribution of client pairs, serving as a benchmark for evaluating model similarity measurements. $LrCos$, depicted in Fig. 2(d), closely aligns with the KL representations, whereas L2 distance (Fig. 2(c)) inadequately captures this similarity. Furthermore, dimensionality reduction significantly reduces the computational burden for similarity calculations. In our settings, the original model dimension is reduced from over $4,700,000$ to $D=50$. Sparse updates to the dimensionality reduction matrix guarantee that the additional computational cost is negligible compared to the saved computation cost, especially in large-scale scenarios.

\subsection{Re-clustering}

FedAC employs a EM-like algorithm for periodic clustering updates. In each E-step, clients are reassigned to nearby clusters based on the $LrCos$ of their personalized models and cluster centroids:

\begin{equation}
R_{i,k} = \begin{cases}
1, & k = \underset{j}{\arg\min} LrCos( \omega_i , \Omega_j ) \\
0,& \textrm{else} \\
\end{cases}
.
\end{equation}

In the M-step, the server aggregates the models within each cluster to obtain the cluster centroids:

\begin{equation}
\Omega_k = \frac{1}{\sum_{i=1}^m R_{i,k}}\sum_{i=1}^m R_{i,k} \omega_i.
\end{equation}

\subsection{Cluster Number Tuning}

In complex heterogeneous environments, manually determining the optimal cluster count $K$ as a predefined hyper-parameter poses challenges for CFL methods. \textsf{FedAC} provides a novel approach to address this by dynamically fine-tuning the value of $K$ until optimal during the training process.

The Cluster Number Tuning (CNT) module in FedAC (Algorithm 3) dynamically assesses model distances within and between clusters (denoted as ${Dist_{intra}}$ and ${Dist_{inter}}$), determining cluster-splitting or merging decisions. We empirically define the ratio $G_c=\frac{{Dist_{intra}}}{{Dist_{inter}}} $ to characterize the clustering granularity and constrain it within a reasonable range. A judicious $G_c$ range prevents insufficient collaboration within a cluster (when $G_c$ is too small) and moderates excessive clustering effects (when $G_c$ is too large).

\begin{algorithm}[H]
\caption{Cluster Number Tuning (CNT)}
\label{alg:CNT}
\textbf{Input}: $ \{ \omega_i \}_{i \in [m]} $, $ \{ \Omega_k \}_{k \in [K]} $\\
\textbf{Parameter}: lower and upper threshold, $a$ and $ b$, of $ G_c $

\begin{algorithmic}[1] 
\FOR{each cluster $k\in [K]$}
\STATE $Dist_{intra}^k = \frac{1}{\sum_{i=1}^m R_{i,k}}\sum_{i=1}^m R_{i,k} || \omega_i - \Omega_k ||^2_2$
\STATE $Dist_{inter}^k = \frac{1}{K-1} \sum_{j=1}^K || \Omega_k - \Omega_j ||^2_2$
\STATE $G_c^k=\frac{{Dist_{intra}^k}}{{Dist_{inter}^k}} $
\IF {$G_c^k < a$}
\STATE Merge cluster $k$ into the closest cluster.
\ELSIF{$G_c^k > b$}
\STATE Divide cluster $k$ into $2$ clusters.
\ENDIF
\ENDFOR
\end{algorithmic}
\end{algorithm}

Drawing from this intuition, the CNT module empirically divides clusters with large $G_c$ values into two new clusters by initializing two new centers and conducting one re-clustering iteration, thereby increasing the total cluster count $K$. Conversely, clusters with excessively small $G_c$ values are merged into the nearest cluster, resulting in a reduction of $K$. Subsequent experiments validate the efficacy of this empirical approach in optimizing $K$ to its optimal value.

\section{Experiments}
\label{fifth:exp}

\subsection{Experimental Setup}
We conduct experiments with two authentic datasets: CIFAR-10 \cite{CIFAR10} and CIFAR-100 \cite{CIFAR10}. Consistent with previous research \cite{CGPFL}, we employed the Dirichlet distribution with $\alpha$ set to $0.1$ and the pathological distribution (where $n$ denotes the number of labels per client) to introduce high heterogeneity. The quantity of instances per client was randomly assigned, spanning from $50$ to $350$ for both datasets. FedAC and the baseline models were implemented using Python 3.7 and PyTorch 1.12.1. The training process took place on a compute server equipped with an NVIDIA RTX 3080 GPU, Intel i9-10900K CPU, 64 GB RAM, and a 2TB SSD. The experiments involved convnets with $100$ clients, sampling $25$ per communication round. These convnets consist of two convolution layers and three fully connected layers, utilizing the final fully connected layer for decision-making and the preceding layers for embedding. The SGD optimizer is employed with a learning rate set to $0.01$. The experimental setup includes a batch size of $32$, local iterations of $5$, and communication rounds of $1000$, unless specified otherwise.

\subsection{Overall Performance}

We conducted a comparative analysis, benchmarking \textsf{FedAC} against seven prominent SOTA methodologies in the field. These methodologies are: (1) \textsf{FedAvg} \cite{FedAvg}, (2) \textsf{FedPer} \cite{FedPer}, (3) \textsf{FeSEM} \cite{FeSEM}, (4) \textsf{FedGroup} \cite{FedGroup}, (5) \textsf{FL+HC} \cite{FLHC}, (6) \textsf{CGPFL} \cite{CGPFL}, and (7) \textsf{IFCA} \cite{IFCA}. Table 1 displays the improved performance of \textsf{FedAC} compared to benchmarks in common heterogeneous scenarios. CFL methods determine their respective optimal cluster numbers $K$. It's important to highlight that \textsf{FeSEM}, \textsf{FedGroup}, \textsf{FL+HC}, and \textsf{IFCA} employ a single shared model for each cluster, whereas \textsf{FedPer}, \textsf{CGPFL}, and the proposed \textsf{FedAC} maintain a personalized model for each client. In contrast to the Dirichlet distribution, the pathological distribution enhances the CFL method's performance by naturally leading certain clients to acquire identical label categories.

\begin{table}[t]
\centering
  \begin{tabular}{lSSSS}
    \toprule
    \multirow{2}{*}{\textbf{Method}} &
    \multicolumn{2}{c}{CIFAR10 Test Accuracy (\%)} &
    \multicolumn{2}{c}{CIFAR100 Test Accuracy (\%)} \\
    & {$\alpha=0.1$} & {$n=3$} & {$\alpha=0.1$} & {$n=8$} \\
      \midrule
    {FedAvg} & {64.75$\pm$1.21}  & {62.73$\pm$0.42}  & {35.36$\pm$0.92} & {33.33$\pm$0.42} \\
    {FedPer} & {70.84$\pm$1.44} & {79.36$\pm$0.54} & {43.46$\pm$1.27} & {62.10$\pm$0.63}\\
    {FeSEM} & {65.65$\pm$1.28} & {76.46$\pm$0.63}  & {31.03$\pm$0.80} & {52.15$\pm$2.45}\\
    {FedGroup} & {67.38$\pm$1.34} & {76.77$\pm$1.00} & {33.26$\pm$1.01} & {57.02$\pm$0.84}  \\
    {FL+HC} & {67.80$\pm$0.84} & {80.22$\pm$0.68} & {34.19$\pm$1.33} & {57.99$\pm$0.66}\\
    {CGPFL} & {71.19$\pm$0.93} & {79.42$\pm$0.46}& {41.38$\pm$0.69} & {60.26$\pm$0.94}\\
    {IFCA} & {73.06$\pm$0.91} & {80.54$\pm$0.74} & {38.61$\pm$0.77} & {60.18$\pm$0.76} \\
    {\textbf{FedAC}} & {\textbf{74.88$\pm$0.65}} & {\textbf{81.29$\pm$0.61}} & {\textbf{51.28$\pm$0.35}} & {\textbf{64.53$\pm$0.34}} \\
    \bottomrule
  \end{tabular}
  \caption{FedAC's performance (mean and standard deviation) is compared to baseline methods in non-IID settings. The best is highlighted in bold.}
\end{table}

\begin{figure*}[htbp]
\centering
\begin{subfigure}{0.235\textwidth}
    \includegraphics[width=\textwidth]{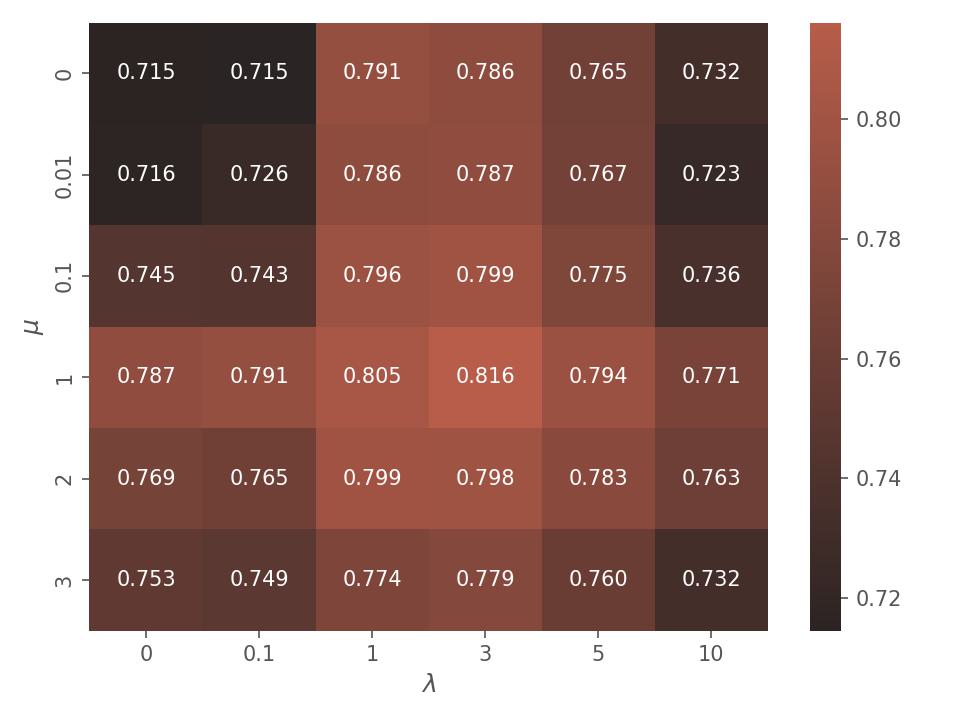}
    \caption{CIFAR-10.}
    \label{fig:changemulamcifar10}
\end{subfigure}
\hfill
\begin{subfigure}{0.235\textwidth}
    \includegraphics[width=\textwidth]{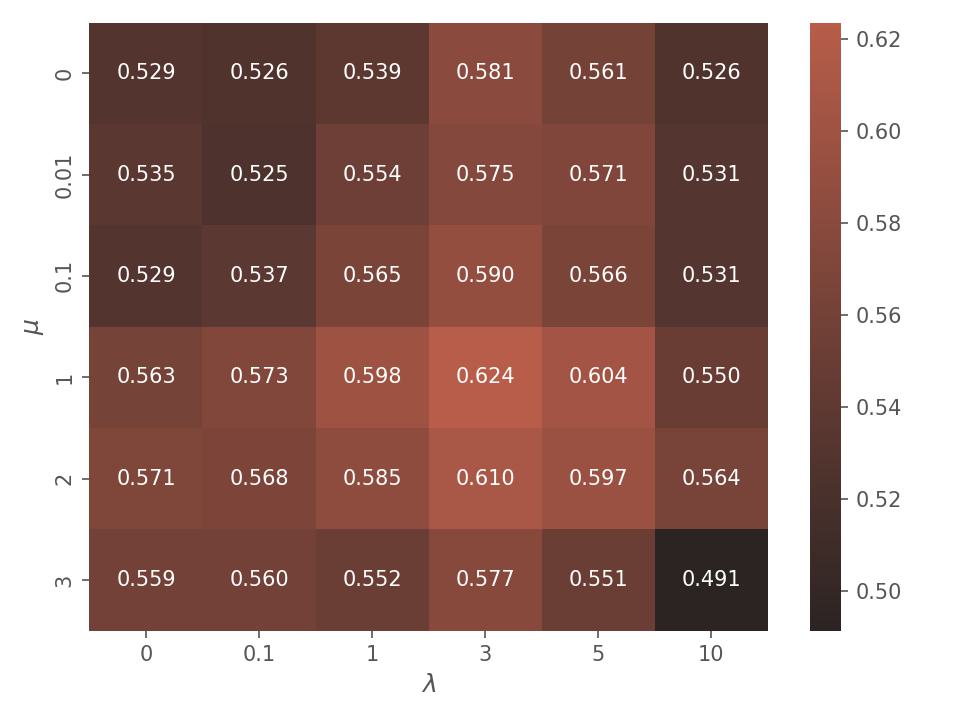}
    \caption{CIFAR-100.}
    \label{fig:changemulamcifar100}
\end{subfigure}
\hfill
\begin{subfigure}{0.235\textwidth}
    \includegraphics[width=\textwidth]{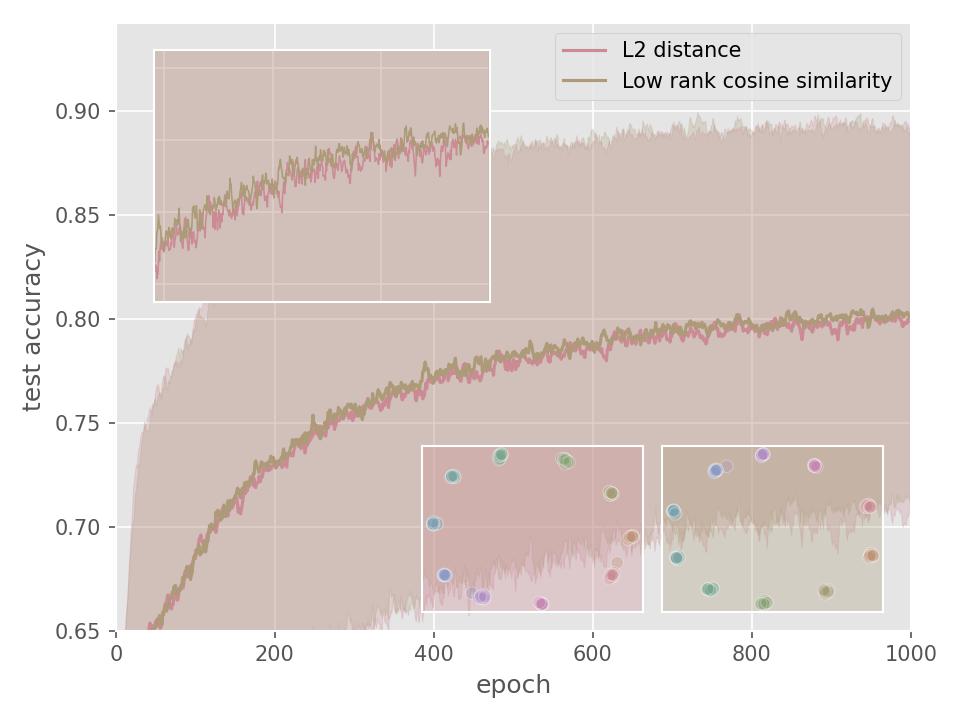}
    \caption{CIFAR-10.}
    \label{fig:l2lrcifar10}
\end{subfigure}
\hfill
\begin{subfigure}{0.235\textwidth}
    \includegraphics[width=\textwidth]{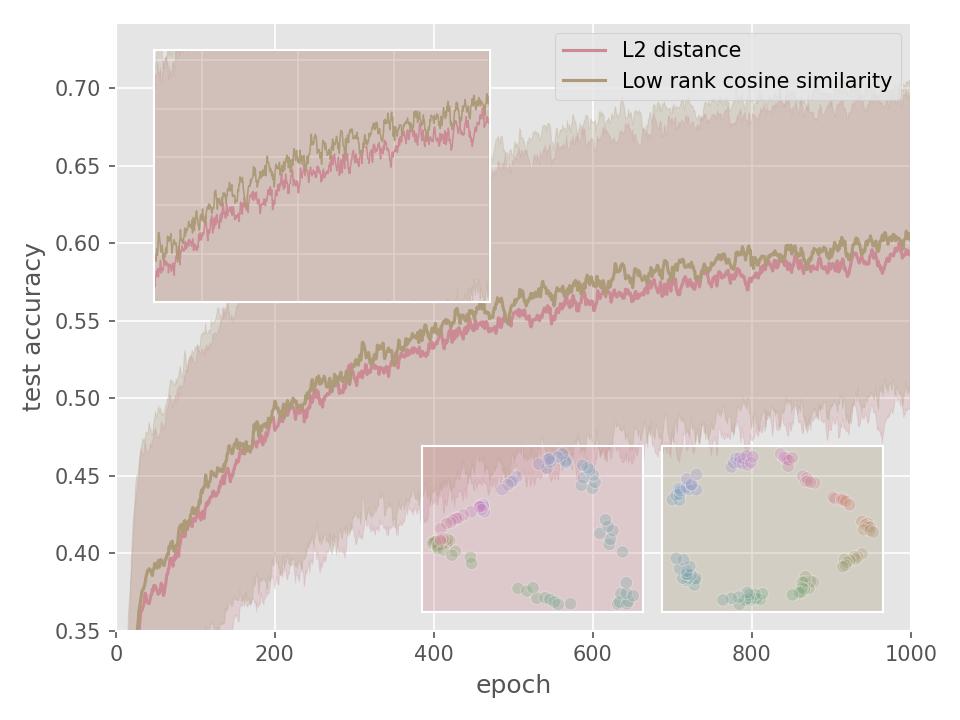}
    \caption{CIFAR-100.}
    \label{fig:l2lrcifar100}
\end{subfigure}

\caption{Ablation experiments were conducted to assess FedAC's performance across different parameter configurations and model similarity measurements.}
\label{fig:changemulamfigures}
\end{figure*}

\textsf{IFCA} demonstrates commendable performance but incurs substantial computational and communication costs. \textsf{CGPFL} preserves personalized models for clients during clustering, yielding enhanced performance in heterogeneous conditions. However, its knowledge sharing is confined to within clusters. \textsf{FedAC} attains optimal performance by amalgamating personalized models, proficient cluster partitioning, and simultaneous coordination of global knowledge sharing.

\begin{figure}[t]
\centering

\begin{subfigure}{0.235\textwidth}
    \includegraphics[width=\textwidth]{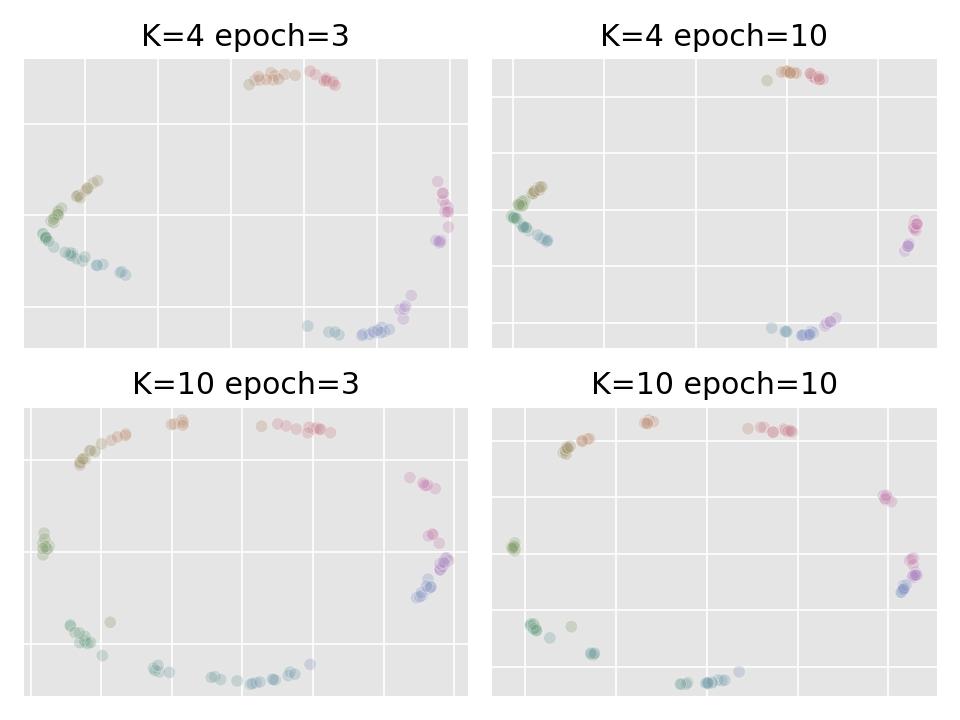}
    \caption{FeSEM.}
    \label{fig:cntchangeone}
\end{subfigure}
\hfill
\begin{subfigure}{0.235\textwidth}
    \includegraphics[width=\textwidth]{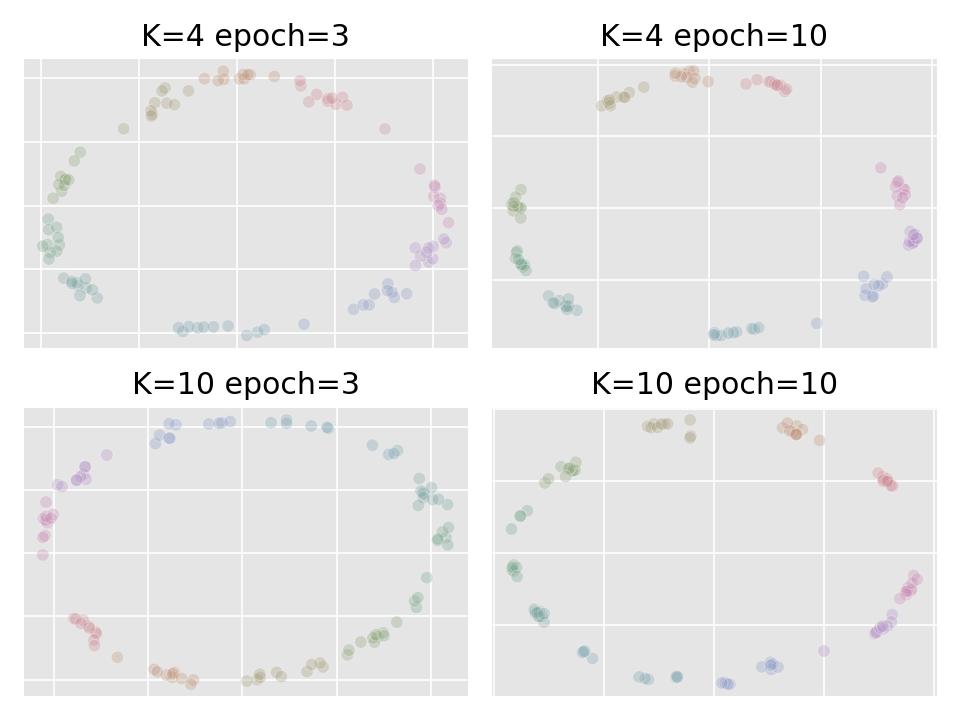}
    \caption{FedAC.}
    \label{fig:cntchangegroup}
\end{subfigure}
\hfill
\begin{subfigure}{0.235\textwidth}    \includegraphics[width=\textwidth]{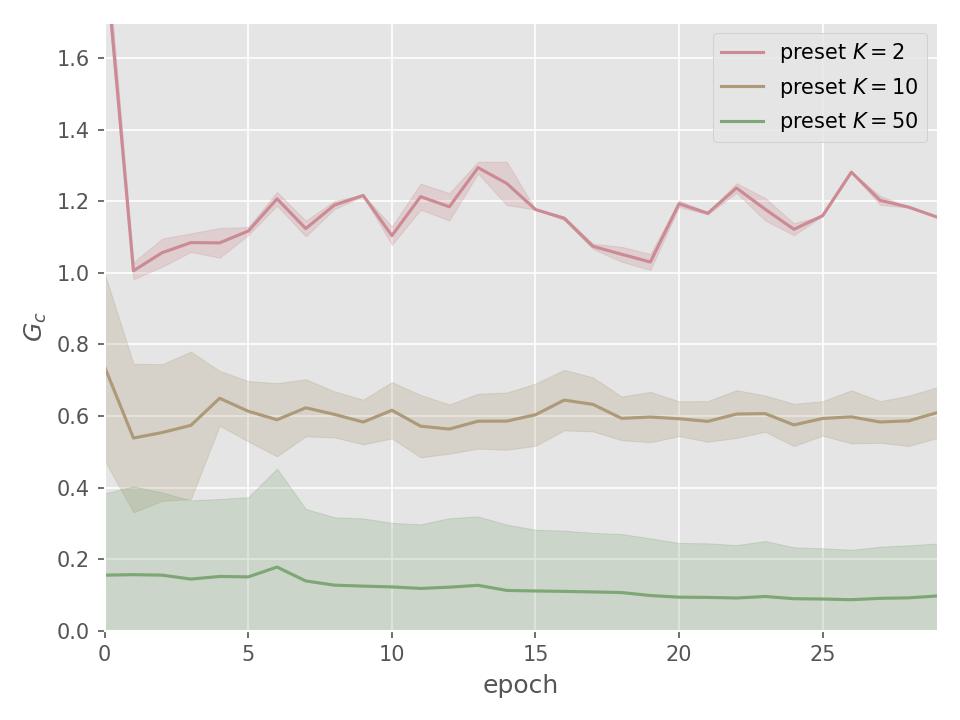}
    \caption{Without CNT.}
    \label{fig:first}
\end{subfigure}
\hfill
\begin{subfigure}{0.235\textwidth}
    \includegraphics[width=\textwidth]{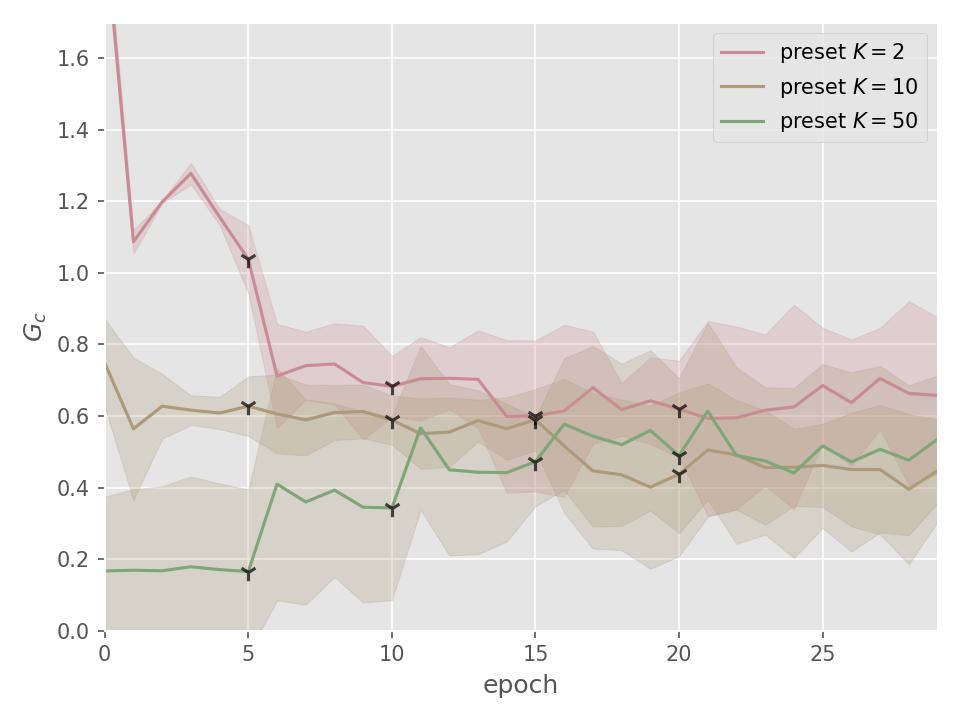}
    \caption{With CNT.}
    \label{fig:second}
\end{subfigure}

\caption{Client models, trained with \textsf{FeSEM} and \textsf{FedAC}, are visualized by projecting them onto a plane in (a) and (b). Clients with similar data distributions are represented by similar colors. In (c), the mean and variance of clusters' $G_c$ are shown for various predefined $K$ values (with $K=10$ as the optimal value), while subplot (d) depicts the CNT module fine-tuning the cluster number by adjusting $G_c$ (marked in black).}
\label{fig:cnteffect}
\end{figure}

\subsection{Global and Intra-Cluster Trade-Offs}

\textsf{FedAC} incorporates regularizing terms for cluster center and global embedding into local model training, fostering alignment between local and global understanding. In Eq. 3, $\mu$ regulates the intensity of cluster center model regularization, while $\lambda$ governs the strength of global embedding regularization. Larger values of these parameters signify a greater reliance on intra-cluster or global information in local updates. Figures 4(a) and 4(b) validate the effectiveness of employing two regularization terms to balance intra-cluster and global information across experimental datasets. Achieving optimal FedAC performance necessitates meticulous adjustment of these regulatory intensities.

\subsection{Improvement in Similarity Measurement}

Following the explanation of the $LrCos$ metric in \cref{subsec:lrcos}, ablation experiments were conducted by substituting $LrCos$ with L2 distance, as depicted in Fig. 4(c) and (d). The utilization of $LrCos$ leads to a more sensible client partitioning and contributes to enhanced model testing accuracy, especially evident in more intricate tasks such as CIFAR-100.

\subsection{Clustering Robustness}

Here, we visualize the clustering for online CFL approaches that lack personalized client models (\emph{e.g.}, \textsf{FeSEM}) and those that incorporate them (\emph{e.g.}, \textsf{FedAC}). The former, which results in swift convergence of models within clusters, contradicts the purpose of re-clustering in online CFL, thereby reducing the system's robustness, as clients within the same cluster initiate updates from identical positions each round (see Fig. 4(a)). \textsf{FedAC}, depicted in Fig. 4(b), addresses this by introducing a soft regularization term ($L_{\text{intra}}$ in Eq. 3) to the personalized models. This also improves the system's resilience to variations in cluster number settings and initializations of cluster centers. Additionally, as shown in Fig. 4(c) and (d), we set the range of the system's $G_c$ to $(0.2, 0.8)$. The CNT module automates the adjustment of $G_c$ to effectively fine-tune to the optimal $K$.

\section{Conclusion}
\label{sixth:conclu}

This paper proposes a clustered federated learning framework that integrates global and intra-cluster knowledge through neural network decomposition. It employs a low-rank cosine similarity for efficient and economical client clustering. Additionally, an integrated module facilitates adaptive cluster tuning to determine the optimal count. The framework's effectiveness is rigorously evaluated through experimental analysis.

%
%
%
%

\bibliography{main} 
\bibliographystyle{IEEEtran}




\end{document}